\documentclass[submission]{eptcs}
\providecommand{\event}{ICLP Doctoral Consortium 2019} % Name of the event you are submitting to
\usepackage{breakurl}  % Not needed if you use pdflatex only.
\usepackage{graphicx}
\usepackage{multirow}
\title{Conversational AI : Open Domain Question Answering and Commonsense Reasoning}
\author{Kinjal Basu
\institute{Erik Jonsson School of Engineering and Computer Science}
\institute{The University of Texas at Dallas \\
Richardson, Texas}
\email{kinjal.basu@utdallas.edu}
}
\def\titlerunning{Conversational AI}
\def\authorrunning{Kinjal Basu}
\begin{document}
\maketitle

\begin{abstract}
Our research is focused on making a human-like question answering system which can answer rationally. The distinguishing characteristic of our approach is that it will use automated common sense reasoning to truly ``understand'' dialogues, allowing it to converse like a human. Humans often make many assumptions during conversations. We infer facts not told explicitly by using our common sense. Incorporating commonsense knowledge in a question answering system will simply make it more robust.

\end{abstract}

\section{Introduction}

The goal of Artificial Intelligence (AI) is to create intelligent systems that can simulate human-like thinking and reasoning process. An intelligent system must be capable of performing automated reasoning as well as responding to the changing environment (for example, changing knowledge). To exhibit such an intelligent behavior, a machine needs to \textit{understand} its environment as well be able to interact with it to achieve certain goals. For acting rationally, a machine must be able to obtain information and understand it. Knowledge Representation (KR) is an important step of automated reasoning, where the knowledge about the world is represented in a way such that a machine can understand and process. Also, it must be able to accommodate the changes about the world (i.e., the new or updated knowledge). Using the generated knowledge base about the world, an intelligent system should be able to do complex tasks like question-answering (QA), summarization, medical reasoning and many more.

Our research is mainly focused on Conversational AI that deals with open domain question answering and commonsense reasoning. Conversational AI has been an active area of research, starting from a rule-based system, such as ELIZA \cite{weizenbaum1966eliza} and PARRY \cite{colby1971artificial}, to the recent open domain, data driven conversational agents like Apple's Siri, Google Assistant or Amazon's Alexa. Early rule-based chat-bots were based on just syntax analysis, while the main challenge of modern machine learning based agents is the lack of ``understanding" of the conversation. These current systems learn patterns in data from large corpora and compute a response without understanding the conversation. Due to lack of ``understanding" of the content, these bots' capabilities are very limited. So, our goal is to create a realistic agent that will ``understand'' the world using commonsense and provide rational answers of any question asked about the world. Like human, this logical agents will also be able to provide justification for every answer.

A realistic agent should be able to understand and reason like a human. In human to human conversations, we do not always tell every detail, we expect the listener to fill gaps through their commonsense knowledge. Also, our thinking process is flexible and non-monotonic in nature, which means \textit{ ``what we believe today may become false in the future with new knowledge"}. We can model this human thinking process with (i) default rules, (ii) exceptions to defaults, and (iii) preferences over multiple defaults \cite{gelfond2014knowledge}. Much of human knowledge consists of default rules, for example, the rule: \textit{Normally, birds fly}. However, there are exceptions to defaults, for example, \textit{penguins are exceptional birds that do not fly}. Reasoning with default rules is non-monotonic, as a conclusion drawn using a default rule may have to be withdrawn if more knowledge becomes available and the exceptional case applies. For example, if we are told that Tweety is a bird, we will conclude it flies. Later, knowing that Tweety is a penguin will cause us to withdraw our earlier conclusion.

\section{Background}
The research in the field of conversational AI started after Alan Turing theorized the Turing Test in 1950. In the year 1966,  Joseph Weizenbaum introduced ELIZA \cite{weizenbaum1966eliza} to the world, which is one of the first chat-bot. Although she was able to fool some users by showing that they are actually talking to a human, ELIZA was not able to clear the Turing Test. ELIZA gives an illusion of understanding, but it does not have any built in framework for contextualizing events. Later on, psychiatrist Kenneth Colby implemented PARRY \cite{colby1971artificial} in 1972 and this is the first chat-bot to pass the Turing Test. After this the evaluation of chat-bot accelerates and also we notice changes in the approaches. Gradually chat-bots started moving from rule-based approaches to data-driven approaches (like machine learning). In 2006, the new era started when IBM came up with ``Watson'' \cite{ferrucci2010building}, who bit former jeopardy champion. In the coming days ``Watson'' started becoming bigger and better, and all other four software giants - Apple, Google, Microsoft and Amazon brought up their chat-bots - Siri, Google-Assistant, Cortana and Alexa respectively.

All these data driven models are very powerful in identifying patterns in data to craft answer without understanding it. So, logical approaches (such as Answer Set Programming) are much better for reasoning and providing rational answers with proper justification. Answer set programming (ASP) \cite{gelfond2014knowledge}, based on the logic programming paradigm, is an elegant method to represent defaults, exceptions and preferences. 
There are comprehensive, well known, implementations of ASP such as CLINGO \cite{DBLP:journals/tplp/GebserKKS19} and DLV \cite{adrian2018asp}. But most of them are based on grounding the logic program and then using a SAT solver to compute a model. These implementations are not scalable (grounding produces an exponential blowup in program size) as well as have other issues that make them impractical for knowledge-representation applications \cite{arias2018constraint}. To overcome these problems, our lab has designed query-driven (goal-directed) implementations of ASP that can handle predicates as well. This implementation, called s(ASP) \cite{arias2018constraint}, has been applied to develop many practical systems that emulates human expertise. For example, it has been used to automate the treatment of congestive heart failure patients, where it outperforms physicians \cite{chen2018ai} (the system automates treatment guidelines from the American College of Cardiology; it can outperform physicians because it acts as an unerring human, while physicians make human errors).

The s(ASP) system has also been used to develop the CASPR open-domain textual question answering system \cite{pendharkar2019asp}. The CASPR system translates a text passage into an answer set program. It then augments it with relevant common sense knowledge imported from resources such as WordNet, representing the imported knowledge as default rules, exceptions, and preferences in ASP. Natural language questions are translated into a query in ASP syntax that is executed against the ASP knowledge-base assembled in the previous step. The execution is done using our s(ASP) query-driven system. The CASPR system can answer questions for passages in the SQuAD data set \cite{rajpurkar2016squad}. Our work  is to enhance the CASPR system with deeper semantic understanding, so that it can handle large textual passages and answer complex questions.

\section{Research Goal}

Our vision is to build an open domain question answering (QA) system that can really “understand” the textual knowledge as well as the context of the text, just like humans do. Most of the current open domain QA systems are based on technologies of information-retrieval (IR) and machine learning (ML). They do use reasoning capabilities, but the reasoning capability is rather shallow and not central to the system. Thus, the answers of most existing QA systems appear unnatural. Our goal is to use (automated) commonsense reasoning as the primary building block of our system, which will be able to answer more logically. Thus, our system will reason and function in a manner similar to humans, and the answers will be more natural. Most of the day-to-day human thinking process can be modeled with (i) default rules, (ii) exceptions to these defaults, and (iii) preferences in presence of multiple defaults. Our system will emulate this human style thinking via modeling of defaults, exceptions and preferences using Answer Set Programming (ASP). 

Broadly, our research goal can be divided into two parts - \textit{textual knowledge understanding} and  \textit{reasoning}.
\subsection{Textual Knowledge Understanding} 

Textual knowledge can come from speech (i.e., speech converted to text using a speech-to-text system), textual passage, image/video summary or a natural language question. Understanding these textual knowledge require semantic understanding and background common sense knowledge. Based on the context the commonsense knowledge will be fetched from various sources (such as WordNet \cite{miller1995wordnet}, ConceptNet \cite{liu2004conceptnet}, Microsoft Concept Graph \cite{yu2016understanding}) and augmented with the textual knowledge base. Our goal is to represent all these knowledge together using facts and default rules with exception in ASP syntax. Natural language questions are converted into an ASP query, which is compatible with the knowledge generated. 

To get deeper understanding of the text we need to parse the text semantically and extract semantic relations. For example , if we have a phrase \textit{``...engine of the train broke down... ''}, then we must extract the information that the \textit{``engine''} is a \textit{``part-of''} the \textit{``train''}.

Natural languages are very ambiguous in nature and that is the biggest challenge in our research. One sentence can be written in many different ways keeping the meaning same, whereas, changing the structure of a sentence with the same words changes the meaning. So, before going to fetch the commonsense knowledge for the background assumption, we need to have basic ambiguity resolutions (for example, word sense disambiguation (WSD), Homonym resolution etc. ). Otherwise, the system may not be able to fetch the relevant information. For example, in a context of `Computer Science', if the word - \textit{``tree''} appears then the system should understands that it is a \textit{``data-structure''} not a \textit{``plant''} or a \textit{``person's name''}.

\subsection{Reasoning}

All the knowledge is represented in ASP syntax and s(ASP) engine will be used to generate possible answer sets. The idea is to reason like human that is non-monotonic in nature. The knowledge will be represented using default rules (as mentioned earlier), so that it can handle new knowledge or modified knowledge of the world. We also assume the world is closed, that means, we assume,  what is not currently known to be true, as false. Negation as Failure (NAF) is used to make the conclusion when the information is absent. This type of negation is used to conclude about default rules and assume defaults to be true in case of absence of enough information. As an example, consider the following example where we state that if we are not able to prove that q(a) succeeds then p(a) succeeds:
\begin{center}
{\small{
\frenchspacing\texttt{p(a) \textrm{:-}  not q(a).}}}\\
\end{center}
So, in the above rule we assumed that p(a) has succeeded based on the absence of information about q(a).

Default reasoning is very useful in modelling human reasoning as we can draw conclusions even in the absence of information by defaulting to the default rule. Default reasoning thus plays an important role in common-sense reasoning and understanding. In case of ASP, a default d stated as “Normally elements of class C have property P” is represented as the following rule:

\begin{center}
{\small{
\frenchspacing\texttt{p(X) \textrm{:-}  c(X), not ab(d(X)), not -p(X).}}}\\
\end{center}
Here, \small{\frenchspacing\texttt{ab(d(X))}} can be read as ``X is abnormal with respect to the default assumption d'' and \small{\frenchspacing\texttt{not p(X)}}  can be read as ``We cannot successfully prove that \small{\frenchspacing\texttt{p(X)}} is false'' or `` \small{\frenchspacing\texttt{p(X)}} may be true''. Default reasoning uses two kinds of exceptions viz strong exceptions and weak exceptions. Weak exception makes the default inapplicable and stop the agent from making a default conclusion. For example, in the above-mentioned default rule we can apply a weak exception \small{\frenchspacing\texttt{ e(X)}} by adding the following rule to the program

\begin{center}
{\small{
\frenchspacing\texttt{ab(d(X)) \textrm{:-}  not -e(X).}}}\\
\end{center}
The exception states that X may not be applicable to d if \small{\frenchspacing\texttt{e(X)}} may be true. Similarly, Strong
Exceptions refute the defaults conclusion by allowing the agent to derive the opposite to be true.
This can be demonstrated by adding the following rule to the program

\begin{center}
{\small{
\frenchspacing\texttt{-p(X) \textrm{:-}  e(X).}}}\\
\end{center}
The above rule states that \small{\frenchspacing\texttt{p(X)}} is false if \small{\frenchspacing\texttt{e(X)}} succeeds, which allows us to defeat d’s conclusion that normally class C elements have the property P.

\section{Current Status of the Research}

The goal directed ASP engine - s(ASP)\cite{arias2018constraint} has been developed in our lab which is available for download (\textit{\url{https://sourceforge.net/projects/sasp-system/}}). Also our lab developed the CASPR system (for open domain question answering), which uses s(ASP) for reasoning. We have developed an unique way to extract semantic relations from text passage using default reasoning (discussed earlier). Those semantic relations represents more deeper understanding of a text passage. Also we have incorporated Microsoft Concept Graph \cite{yu2016understanding} to CASPR for generating relevant commonsense knowledge. Our current goal is to enhance the CASPR system to handle complex questions and apply the system in different applications (like, image question answering).

\section{Preliminary Result}
CASPR system \cite{pendharkar2019asp} was tested on Stanford's SQuAD dataset and the results are very promising. The SQuAD Dataset \cite{rajpurkar2016squad} contains more than 100,000 reading comprehensions along with questions and answers for those reading passages. SQuAD dataset uses the top 500+ articles from the English Wikipedia. These articles are then divided into paragraphs. The Dev Set v1.1 of the SQuAD Dataset has been used to obtain comprehension passages for building a prototype for the proposed approach. This dataset has around 48 different articles with each article having around 50 paragraphs each. Out of the 48 different articles in the SQuAD dev set, 20 articles were chosen from different domains to help build the CASPR system. Using the 20 different articles mentioned above, the ASP program was generated on one paragraph from each article. Then, ASP queries were generated for all the questions in the dataset for these paragraphs. The results show the percentage of questions for which the answer generated from the ASP solver was present in the list of answers specified for the question in the SQuAD dataset. This result can be viewed in terms of the Table \ref{tab:result_label}. The result shows that approximately 80\% of the questions can be answered. This shows that most of the knowledge, if not all, has been captured successfully in the ASP program generated for the passage. The ASP queries generated for the questions are very similar to the original question and convey the same meaning.

\begin{table}[ht]
    \centering
\begin{tabular}{|c||c||c|c|c|}
 \hline 
 \multirow{2}{*}{No}  & \multirow{2}{*}{Article}  & \multicolumn{2}{c|}{Result} & \multirow{2}{*}{Percent} \\ 
   \cline{3-4} 
    &  & Correct & Question Count &  \\ 
   \hline
   1 & ABC & 5 & 5 & 100.00\\
   \hline
   2 & Amazon Rainforest & 12 & 14 & 85.71\\
   \hline
   3 & Apollo & 4 & 5 & 80.00\\
   \hline
   4 & Chloroplasts & 4 & 5 & 80.00\\
   \hline
   5 & Computational Complexity & 3 & 3 & 100.00\\
   \hline
   6 & Ctenophora & 9 & 12 & 75.00\\
   \hline
   7 & European Union Law & 13 & 13 & 100.00\\
   \hline
   8 & Genghis Khan & 3 & 5 & 60.00\\
   \hline
   9 & Geology & 4 & 5 & 80.00\\
   \hline
   10 & Immune System & 13 & 15 & 86.67\\
   \hline
   11 &  Kenya & 5 & 5 & 100.00\\
   \hline 
   12 &  Martin Luther & 2 & 5 & 40.00\\
   \hline
   13 & Nikola Tesla & 6 & 7 & 85.71\\
   \hline
   14 & Normans & 4 & 5 & 80.00\\
   \hline
    15 & Oxygen & 8 & 15 & 53.33\\
   \hline
    16 & Rhine & 5 & 8 & 62.50\\
   \hline
    17 & Southern California & 3 & 5 & 60.00\\
   \hline
    18 & Steam Engine & 4 & 5 & 80.00\\
   \hline
    19 & Super Bowl 50 & 25 & 29 & 86.21\\
   \hline
    20 & Warsaw & 3 & 5 & 60.00\\
   \hline
   \multicolumn{2}{|c||}{Total} & 135 & 171 & 78.95 \\
   \hline
   \multicolumn{2}{|c||}{Average Result} & \multicolumn{3}{c|}{77.76} \\
   \hline
\end{tabular}     \caption{Results of Question Answering}
    \label{tab:result_label}
\end{table}

\section{Open Issues and Expected Achievements}
In our approach we have two major challenges - (i) \textit{Sentence Parsing} and (ii) \textit{Commonsense Knowledge Generation}. Both of them are equally important to be solved. In the following sections we will discuss about the open issues.

\subsection{Sentence Parsing} In this research we are dealing with lots of natural language sentences (mainly in English). So, parsing the sentences syntactically and semantically are very important to know the meaning of the whole sentences and the context. Currently we are using Stanford Parser \cite{manning2014stanford} to tag the Parts-of-Speech (POS) and get the dependency graph. Then we use our default rules to fetch the semantic relations of the text. Eventually we will create a semantic parser to get deeper semantic relations of the sentence.

\subsection{Commonsense Knowledge Generation} The commonsense knowledge are the backbone of this research. In a human conversation we assume many untold things. These back ground assumptions are actually the commonsense knowledge we gathered from our day to day life. There are many projects like Cyc \cite{lenat1995cyc} , ConceptNet \cite{liu2004conceptnet}, FrameNet \cite{baker1998berkeley}, Microsoft Concept Graph \cite{yu2016understanding} etc. Most of these data are crowd-sourced or collected as a flat knowledge. Also, when we retrieve these data, we do not have any control over the commonsense knowledge generated in the background. Instead of generating relevant information, it may generate bunch of unnecessary information, which we may not require. So, our vision is to represent commonsense knowledge in a more human-like way.

\section{Conclusion}
We are standing in an era where the world is gradually moving from touch-screen systems to voice-controlled devices. In the current time, the usage of voice driven system is increasing very fast. Already we have started using voice driven digital assistant from the big four systems - Apple’s Siri, Microsoft’s Cortana, Amazon’s Alexa and Google’s Assistant. These systems are very user friendly as it can handle natural languages and also it does not need any expertise to use them.

Our long term goal is to create a social-bot, that can communicate with human as a friend. It will not be an information retrieval bot, but beyond that it can understand emotion and intention of a human from the conversation using commonsense.

\nocite{*}
\bibliographystyle{eptcs}
\bibliography{generic}
\end{document}